\documentclass{article}

\usepackage{PRIMEarxiv}

\usepackage[utf8]{inputenc} 
\usepackage[T1]{fontenc}    
\usepackage{authblk}
\usepackage{hyperref}       
\usepackage{url}            
\usepackage{booktabs}       
\usepackage{amsfonts}       
\usepackage{nicefrac}       
\usepackage{microtype}      
\usepackage{lipsum}
\usepackage{fancyhdr}       
\usepackage{graphicx}       
\graphicspath{{media/}}     
\usepackage{amsmath}
\usepackage{amsthm}  
\usepackage{amssymb}
\usepackage{algpseudocode}
\usepackage{tabularx}
\usepackage{multirow}
\usepackage{bm}
\usepackage[lined,boxed,commentsnumbered, ruled]{algorithm2e}
\usepackage{soul}
\usepackage{color, xcolor}
\usepackage[shortlabels]{enumitem}
\usepackage{subfigure} 
\title{A Split-and-Privatize Framework for Large Language Model Fine-Tuning}
\author[1]{Xicong Shen}
\author[1]{Yang Liu}
\author[1]{Huiqi Liu}
\author[1]{Jue Hong}
\author[1]{Bing Duan}
\author[2]{Zirui Huang}
\author[2]{Yunlong Mao}
\author[1]{Ye Wu}
\author[1]{Di Wu}
\affil[1]{Bytedance, China}
\affil[2]{Nanjing University, China}

\date{} 

\begin{document}
\maketitle

\begin{abstract}
Fine-tuning is a prominent technique to adapt a pre-trained language model to downstream scenarios. In parameter-efficient fine-tuning, only a small subset of modules are trained over the downstream datasets, while leaving the rest of the pre-trained model frozen to save computation resources. In recent years, a popular productization form arises as Model-as-a-Service (MaaS), in which vendors provide abundant pre-trained language models, server resources and core functions, and customers can fine-tune, deploy and invoke their customized model by accessing the one-stop MaaS with their own private dataset. In this paper, we identify the model and data privacy leakage risks in MaaS fine-tuning, and propose a Split-and-Privatize (SAP) framework, which manage to mitigate the privacy issues by adapting the existing split learning architecture. The proposed SAP framework is sufficiently investigated by experiments, and the results indicate that it can enhance the empirical privacy by $62\%$ at the cost of $1\%$ model performance degradation on the Stanford Sentiment Treebank dataset.
\end{abstract}


\section{Introduction}
In recent years, pre-trained language models (PLMs) represented by BERT and GPT have demonstrated powerful text learning capabilities \cite{kenton2019bert, brown2020language} and have been widely used in various fields such as law \cite{xiao2021lawformer, jiang2023legal}, finance \cite{araci2019finbert, arslan2021comparison}, and healthcare \cite{arora2023promise}. To improve the adaptability of a PLM on downstream applications, it is necessary to fine-tune it on datasets related to the downstream tasks. Considering that PLMs contain hundreds of millions of parameters, researchers have proposed several parameter-efficient fine-tuning (PEFT) algorithms to reduce the cost of secondary training \cite{ding2023parameter}, such as LoRA \cite{hu2021lora}, prefix tuning \cite{li2021prefix}, and prompt tuning \cite{lester2021power}. However, many users are unable to independently acquire the PLM and perform fine-tuning due to resource or technical constraints, which has given rise to a new business direction known as model-as-a-service (MaaS). In MaaS, enterprises with ample resources and technical capabilities (called vendors) release PLMs in the form of cloud services and provide customers with a fine-tuning API so that they can customize their own LLM based on private data. 

However, while this solution provides customers with efficient and customizable LLM services, it also carries the risk of privacy leakage. On the one hand, since the pre-training process requires a large amount of computational overhead, the weights of PLMs are typically considered as proprietary assets of vendors and cannot be made public. On the other hand, customers' text data usually contains sensitive information such as identity and asset, so directly transmitting the original data or representations to the vendor may result in serious privacy leaks \cite{pan2020privacy, qu2021natural,song2020information}, which hinders privacy-conscious customers from using the customization service. Therefore, there is an urgent need for a privacy-preserving fine-tuning framework to alleviate privacy concerns and promote the development of customized services for LLM.

Some prior works have ventured into this domain, albeit encountering certain challenges along the way. For example, the work \cite{qu2021natural} proposed a text privatization mechanism based on $d\chi$-privacy, where the consumer perturbs each individual data entry locally before releasing it to the vendor. But it must be acknowledged that implementing text privatization will invariably lead to performance degradation on downstream tasks \cite{qu2021natural, li2023privacy}, involving the delicate trade-off between utility and privacy. In order to protect both parties' privacy and achieve efficient fine-tuning, the work \cite{xiao2023offsite} proposed the offsite-tuning framework, in which the vendor sends a lightweight adapter and a lossy compressed emulator to the customer, and the customer performs fine-tuning and then returns the final adapter.
However, this framework does not consider privacy concerns during the inference phase.

To address the challenges mentioned above, we propose a Split-and-Privatize (SAP) federated fine-tuning framework based on the existing split learning architecture \cite{vepakomma2018split, romanini2021pyvertical, ceballos2020splitnn}. Specifically, the vendor first splits the entire PLM into a bottom model and a top model, and sends the bottom model to the customer while preserving the confidentiality of the majority of the PLM. During fine-tuning, the customer feeds local sensitive data into the bottom model and privatizes the outputs by applying privacy-preserving mechanisms before sending them to the vendor.
Furthermore, in order to improve the utility-privacy trade-off caused by privatization, we propose a contributing-token-identification (CTI) method. By reducing the perturbation to a small number of token representations that are strongly related to the utility task, we significantly improve the utility performance while maintaining a similar level of empirical privacy. In short, the proposed SAP framework serves as a comprehensive solution that simultaneously protects the vendor's model privacy and the customer's data privacy, and provides a more sophisticated  trade-off between utility and privacy.

In order to comprehensively evaluate the performance and security of the SAP framework, we conduct a series of experiments covering multiple natural language processing tasks, including sentiment analysis, topic classification and semantic equivalence judgment. Additionally, we conduct simulated privacy attacks to validate the effectiveness of SAP on protecting data privacy. 
Experimental results indicate that the proposed framework can achieve a good balance between protecting model
privacy and data privacy while maintaining competitive performance.

The remainder of this paper is organized as follows. In Section 2, we introduce some related works and preliminaries. In Section 3, we present the considered problem setting. In Section 4, we formulate the SAP federated fine-tuning framework and propose the CTI method. Followed by it, experimental results are presented in Section 5 to demonstrate the effectiveness of the proposed framework. Finally, some conclusions are given in Section 6.

\section{Related Works and Preliminaries}
\subsection{Related Works}
\subsubsection{Parameter-efﬁcient Fine-tuning}
Since the introduction of the transformer architecture by Google scholar Ashish Vaswani in 2017 \cite{vaswani2017attention}, PLMs based on the deep transformer architectures have opened a new era of natural language processing. Typically, PLM first models natural language in an auto-regressive or sequence-to-sequence manner on a large-scale unsupervised corpus \cite{radford2018improving, raffel2020exploring}, and then specific objectives are introduced to fine-tune the PLM to adapt to downstream tasks. 

As the scale of PLM expands, the full-parameter fine-tuning method becomes impractical due to its growing time and resource consumption. To solve this problem, a batch of PEFT methods including adapter-based tuning and prompt-based tuning have been proposed \cite{ding2023parameter}. The essence of these methods is to update only a small number of the trainable parameters that can be placed anywhere within the PLM. Adapter-based approaches incorporate small-scale adapters into PLMs and only adjust these adapters for model adaptation. For example, LoRA \cite{hu2021lora} injects trainable rank decomposition matrices into each attention layer of the transformer architecture. BitFit \cite{zaken2022bitfit} can achieve performance comparable to full-parameter fine-tuning on small-to-medium datasets by simply updating the bias terms of PLMs. Unlike adapter-based approaches, prompt-based approaches aim to guide the PLM to perform specific tasks by designing and fine-tuning prompts or input instructions. Prompt tuning \cite{lester2021power} prepends a small number of trainable tokens before the input text and optimizes them while keeping the entire model frozen. Although the above PEFT methods can efficiently adapt PLMs to downstream tasks, they primarily focus on centralized scenarios and cannot address the challenges related to model and data privacy in the  LLM customization service.

\subsubsection{Split Learning}
Split learning (SL) is a distributed learning technique that divides the entire model into multiple segments held by different parties, allowing multiple parties to collaboratively train the model without revealing their original data \cite{vepakomma2018split}. In SL, split neural network (SplitNN) is the most commonly used paradigm \cite{romanini2021pyvertical, ceballos2020splitnn}. Speciﬁcally, the entire neural network is split into a top network on the server and multiple bottom networks held by different clients. During training, each client transforms its local input data into intermediate features and sends them to the server. The server first concatenates all the intermediate features and continues to perform forward propagation in the top model to compute the loss. Then it performs backpropagation to compute the gradient and sends the corresponding intermediate-layer gradient to each client so that they can complete the update of their local network segment.

Although SL has the advantage of avoiding the disclosure of raw data, some studies have shown that there is still a potential risk of privacy leakage when clients directly transmit intermediate representations \cite{dosovitskiy2016inverting, he2020attacking}. For instance, the work \cite{he2020attacking} proposed attack methods for both the white-box and black-box scenarios, which can partially recover the original inputs from the transmitted representations. Therefore, SL also needs to be combined with technologies such as differential privacy and homomorphic encryption to enhance privacy protection.

\subsubsection{Privacy-preserving LLM Services}
Existing research on privacy-preserving LLM mainly focuses on centralized learning and addresses concerns about the potential privacy leakage of training data when deploying LLMs publicly, which is referred to as the memory privacy of LLMs \cite{carlini2021extracting, peris2023privacy}. 
For the MaaS scenario, a few of studies effort to propose methods that protect both model privacy and data privacy.
Typically, the vendor keeps the backbone of the PLM confidential at the cloud server and only releases the embedding layer. The customer is then required to send perturbed texts or text representations to the vendor to complete subsequent fine-tuning and inference. The work \cite{lyu2020differentially} proposed a method based on differential privacy (DP) and word dropout, which protects data privacy during the inference phase by randomly dropping some words and adding Gaussian noise to the text representation. The work \cite{lyu2020towards} proposed a local differential privacy (LDP) scheme, where the customer encodes its input embedding vectors in binary and then randomly flip some of the positions to introduce perturbations. Chen et al. \cite{qu2021natural} investigated the impact of applying the $d\chi$-privacy mechanism (a variant of LDP) to BERT fine-tuning on both privacy and utility, and proposed a privacy-adaptive LLM pre-training method, which applies the same perturbations to the pre-training corpus to improve the practicality of the fine-tuned model. Compared with our work, \cite{qu2021natural} requires retraining the PLM based on the designed masked LM objective on the publicly available corpora, which incurs significant computation costs.
Besides, considering the high cost of fine-tuning the entire model on private data, Li et al. \cite{li2023privacy} proposed a privacy-preserving prompt tuning framework called RAPT, where the customer applies text-to-text privatization based on the $d\chi$-privacy locally and the vendor performs prompt tuning on the privatized data. Furthermore, due to the poor performance of prompt fine-tuning directly on the privatized data, the work \cite{li2023privacy} also introduced a token reconstruction task jointly trained with the downstream task, so that the prompt obtained by training can help LLMs learn better task-related representations. In comparison, we propose a more general federated fine-tuning framework SAP. By splitting some encoder blocks to the bottom model instead of just the embedding layer, the SAP framework achieves a better trade-off between model performance and data privacy. Furthermore, compared with the token reconstruction method, the proposed CTI method saves extra resources required for plain token transmission and reconstructing model training.


In addition, Xiao et al. \cite{xiao2023offsite} introduced a novel approach called offsite-tuning for privacy-preserving LLM customization. In this framework, the vendor first sends a lightweight adapter and a lossy compressed PLM simulator to the customer, who then fine-tunes the adapter locally on downstream data with the help of simulator. Finally, the customer returns the fine-tuned adapter, and the vendor incorporates it into the PLM to obtain a fine-tuned LLM. The offsite-tuning framework shares some similarities with our SAP framework, both deploy some simplified modules on the customer and simultaneously consider the model privacy and data privacy. However, offsite-tuning does not consider privacy concerns during the inference phase, while the proposed SAP is equally applicable to inference.


\subsection{Preliminaries}
In this subsection, we introduce some preliminaries about $d\chi$-privacy mechanism used in this paper, which is a variant of LDP  \cite{chatzikokolakis2013broadening}. The specific definition of $d\chi$-privacy is given below.


\newtheorem{definition}{Definition}
\begin{definition}
A randomized mechanism $\mathcal{M}$ satisﬁes $\eta d\chi$-privacy if for any two inputs $x,x^{\prime} \in \mathcal{X}$, 
\begin{equation}   \label{eqn1}                         
\frac{\operatorname{Pr}[M(x)=y]}{\operatorname{Pr}\left[M\left(x^{\prime}\right)=y\right]} \leq e^{\eta d\left(x, x^{\prime}\right)}, \forall y \in \mathcal{Y},
\end{equation}
where $\eta >0$ is a privacy parameter and $d(x,x^{\prime})$ is a distance function.
\end{definition}

Compared with the definition of LDP, $d\chi$-privacy replaces the exponent term on the right side of the inequality \eqref{eqn1} from $\epsilon$ to $\eta d\left(x, x^{\prime}\right)$, so it is a relaxation of LDP. LDP is a strong privacy standard, which requires that any two inputs have similar and indistinguishable output distributions regardless of how different they are. Therefore, the output may not retain enough information of the original input, resulting in severe performance degradation. In contrast, $d\chi$-privacy allows the indistinguishability of the output distributions to be scaled by the distance between inputs, which enables the randomized mechanism to retain more information about input.

The work \cite{feyisetan2020privacy} proposed a text-to-text privatization mechanism that guarantees the $\eta d\chi$-privacy. For any word $w$, the mechanism first computes the embedding vector $\phi(w)$, and then adds appropriate random noise $\bm{n}$ to obtain the perturbed vector $\hat{\phi}(w)=\phi(w)+\bm{n}$, where the probability density of the noise satisfies $p(\bm{n}) \propto \exp (-\eta\|\bm{n}\|)$. Finally the word $w$ is replaced by the word $w^\prime$ closest to $\hat{\phi}(w)$.

\section{Problem Setting}
\subsection{Problem Definition}
In this paper, we focus on the customization of LLM involving two parties, where the vendor holds the complete PLM $\bm{w}$ and abundant server resources, and the customer holds the private labeled dataset $\mathcal{D}:=\{(\bm{x}_{i},y_{i})|i=1,2,\dots,|\mathcal{D}|\}$ and limited computation resources. In order to achieve optimal adaptation on the downstream task, the vendor and customer need to collaboratively fine-tune the PLM, which can be formulated as
\begin{equation}   \label{eqn2}                     
\arg \min_{\bm{\delta}} \mathcal{L}(\bm{w}+\bm{\delta}, \mathcal{D}).
\end{equation}
However, due to privacy constraints, the above fine-tuning process cannot be performed in a centralized manner on one party. Specifically, the privacy constraints include that the vendor cannot share the PLM $\bm{w}$ with the customer, and the customer cannot share the private dataset $\mathcal{D}$ with the vendor. 

To solve the above problem, we propose a privacy-preserving federated fine-tuning framework called  Split-and-Privatize (SAP). As the name suggests, the SAP framework splits the PLM $\bm{w}$ into a top model $\bm{w}_t$ and a bottom model $\bm{w}_b$, which are deployed on the vendor and customer respectively. During fine-tuning, the customer is required to convert its input text into text representations (i.e., the output of the bottom model) and transmit them to the vendor to complete training. In order to enhance privacy protection, the customer is recommended to employ techniques such as differential privacy to perturb text representations, which is referred to as text privatization.

\subsection{Threat Model and Design Goals}
For the threat model, we assume that both participants are honest-but-curious, that is, they always follow the designed SAP framework but are curious about other's private information (i.e. the private data of customer and the model parameters of vendor). The customer might peek at the model architecture and parameters transferred from the vendor, while the vendor may attempt to infer some privacy information from text representations transmitted from the customer, such as the embedding inversion attack \cite{qu2021natural} and the attribute inference attack \cite{song2020information}. A detailed description of the aforementioned attack methods is presented in the experiment settings of Section 5.
Despite the risk of information leakage, we assume no adversary can corrupt the SAP framework.

Under the above threat model, the proposed SAP framework aims to achieve the following goals. Firstly, most parameters of the PLM cannot be disclosed to the customer. Secondly, the SAP framework should ensure that 
it is difficult for the vendor to recover the original input text from the transmitted representations. Lastly, the performance of SAP should not degrade significantly compared to centralized fine-tuning.

\section{Proposed Method}
In this section, we first give an overview of the proposed SAP framework and then introduce two important modules (model split and text privatization) in detail. Following that, a CTI method is proposed to improve the utility-privacy trade-off. Finally, we give a specific algorithm implementation.

\subsection{Split-and-Privatize Framework}

\begin{figure}[t]
\centering
\includegraphics[width=4.5in]{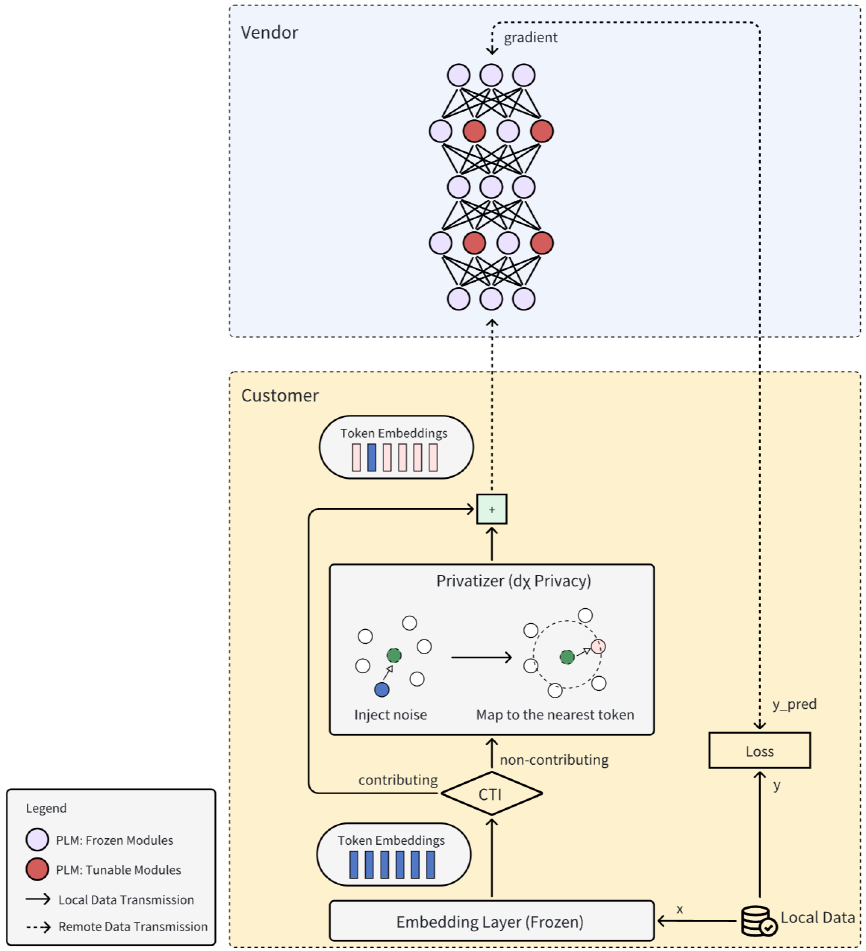}
\caption{An overview of the SAP-CTI algorithm, where the PLM is split into a bottom model (embedding layer) and a top model, and the customer privatizes the embedding vectors before releasing them.}\label{fig1}
\end{figure}



To protect both the vendor's model privacy and the customer's data privacy while achieving LLM customization, we propose the SAP framework based on the existing split learning architecture. Figure 1 shows an implementation example of the SAP framework with CTI. In general, the entire PLM is split into the top model on the vendor and the bottom model on the customer. During fine-tuning, the customer first computes the outputs of the bottom model on local private data, adaptively applies privacy-preserving mechanisms to privatize the text representations and then sends them to the vendor. After receiving the perturbed representations, the vendor proceeds to perform forward propagation in the top model to compute the output of the PLM. Since the sample labels are also held by the customer, in order to update trainable parameters in the PLM such as the Lora module, the vendor needs to send the output to the customer and receive the gradients of the output layer in return.
The following two subsections provide detailed descriptions of the model split and text privatization modules in the SAP framework.


\subsection{Model Split}
In the initialization phase, similar to the idea of SplitNN \cite{romanini2021pyvertical}, the vendor splits the PLM into a bottom model and a top model, and sends the bottom model to the customer. The choice of which layer to split the model is an important option in SAP. 
If the bottom model only has an embedding layer, the customer's computational burden is relatively small, but the vendor can easily recover the input text from the transmitted representations by nearest neighbor search \cite{qu2021natural}. If the bottom model contains more encoder blocks, it becomes more difficult to recover the original text. The work \cite{song2020information} proposed an attack method based on word selection vector optimization and demonstrated that inverting input text from higher layers of deep model is more challenging, as the representations at higher layers are more abstract and generic.
In addition, since the weights of PLM are also important assets, the vendor may request to release as few weights as possible. Therefore, determining the split position of PLM requires a comprehensive consideration of multiple factors.

Besides, the SAP framework can be divided into two cases depending on whether the bottom model is trainable or not. For the case where the bottom model is frozen, the customer computes the representations of all samples immediately after receiving the bottom model, adds perturbations and sends them to the vendor all at once. Therefore, in the subsequent fine-tuning phase, the customer does not need to repeatedly perform forward computations locally. Another implementation is to make the bottom model on the customer also trainable. It is hoped that by altering the parameters of the bottom model during the fine-tuning process, it will be more challenging for the vendor to infer the original input text based on the released text representations. Specifically, during forward propagation, the customer no longer sends  the perturbed representations of all samples to the vendor at once. Instead, in each iteration, the customer randomly selects a batch of samples, computes the output of the bottom model for that batch, and then sends the privatized results to the vendor. In addition, during backpropagation, the vendor needs to return the gradient of the input layer of top model to the customer so that it can update the bottom model.

\subsection{Text Privatization}
Given the bottom model, the customer can obtain the text representation for each sample. However, if plain representations are directly released, the vendor might be able to accurately recover the original input text \cite{song2020information}. Therefore, to protect data privacy, it is necessary for the customer to employ privatization mechanisms to perturb text representations. The proposed SAP framework can be easily combined with various privacy-preserving mechanisms to achieve stronger privacy protection, such as $d\chi$-privacy \cite{chatzikokolakis2013broadening} and probably approximately correct (PAC) privacy \cite{xiao2023pac}. We take the case where the bottom model is a frozen embedding layer as an example to demonstrate how to combine the SAP framework with the privatization mechanism proposed in \cite{feyisetan2020privacy} to guarantee $\eta d\chi$-privacy.

Let $[x_{i}^{1},x_{i}^{2},\dots,x_{i}^{n}]$ represent a sequence of tokens for the input text $\bm{x}_{i}$. 
The customer first obtains the embedding vector $\phi(x_{i}^{j})$ for each token $x_{i}^{j}$ in the sample $\bm{x}_i$ based on the embedding layer. Then independent random noise $\bm{n}$ is added to each embedding vector,
\begin{equation}   \label{eqn3}                     
\hat{\phi}(x_{i}^{j}) = \phi(x_{i}^{j}) + \bm{n}, \quad p(\bm{n}) \propto \exp (-\eta\|\bm{n}\|).
\end{equation}
The specific generation method of noise $\bm{n}$ can be found in Section 2.6 of \cite{feyisetan2020privacy}. Then the perturbed vector is replaced by its nearest neighbor in the embedding space,
\begin{equation}   \label{eqn4}                     
\bar{\phi}(x_{i}^{j})= \arg \min _{\bm{w}_l}\left\|\hat{\phi}(x_{i}^{j})-\bm{w}_l\right\|,
\end{equation}
where $\bm{w}_l$ represents the embedding vector in the embedding space. Finally, the customer sends $|\mathcal{D}|$ perturbed sequences $\bar{\phi}(\bm{x_{i}})=[\bar{\phi}(x_{i}^{1}),\bar{\phi}(x_{i}^{2}),\dots,\bar{\phi}(x_{i}^{n})]$ to the vendor.

For the case where the bottom model is trainable, the customer needs to independently privatize the representations of the current batch of samples in each iteration, and the specific privatization process is consistent with the case where the bottom model is frozen. 



\subsection{Contributing Token Identification}
In order to strengthen the protection of data privacy, the SAP framework introduces the text privatization module. However, fine-tuning PLM on the perturbed representations will inevitably lead to performance degradation on the downstream task, so there is a trade-off between utility and privacy. To improve the utility-privacy trade-off of the SAP framework where the bottom model only contains the embedding layer, we propose a contributing-token-identification (CTI) method. The key rationale of this method is to use statistical analysis to identify the tokens that contribute the most to the utility target in each class of samples, and then reduce the perturbations applied to these specific tokens, aiming to improve utility performance while maintaining a similar level of privacy protection. The following is a detailed description of this method.

In natural language processing, term frequency-inverse document frequency (TF-IDF) \cite{salton1988term} is a metric used to measure the importance of a word to a document in a collection or corpus, which is widely used in information retrieval and text mining. The TF-IDF value is proportional to the frequency of a word appearing in a document, and inversely proportional to the proportion of documents containing this word in the corpus, thereby reducing the impact of common words. Inspired by TF-IDF, we propose a metric that measures the importance of each token in relation to the utility target for text classification tasks. Let $p(t=t_m|y=c)$ represent the frequency of token $t_m$ appearing in the $c$-th class of samples, then the utility importance (UI) of token $t_m$ to class $c$ is defined as
\begin{equation}   \label{eqn5}            
\text{UI}_{mc}=\sum_{c^{\prime},c^{\prime}\neq c}{\ln\frac{p(t=t_m|y=c)}{p(t=t_m|y=c^{\prime})}},
\end{equation}
where $\ln\frac{p(t=t_m|y=c)}{p(t=t_m|y=c^{\prime})}$ can be regarded as the difference between the probability distribution of tokens in the $c$-th class of samples and that in the $c^{\prime}$-th class of samples specifically at token $t_m$. Intuitively, tokens that appear frequently in the $c$-th class of samples while having low frequency in other class of samples will be considered to contribute significantly to distinguishing the $c$-th class from other classes, and thus will be assigned a larger UI value.

For each class of samples, the customer first computes the utility importance of each token in the vocabulary and selects the top-$k$ tokens with the largest UI value as the contributing tokens.
Then, when applying text privatization locally, the customer reduces the perturbation to the embedding vectors of these tokens to enhance utility performance. The experimental results in Section 5.2 indicate that when the contributing tokens account for only $1\%$ of all tokens, the CTI method can significantly enhance the practicality of LLM on the downstream task while maintaining a similar level of privacy protection.

\renewcommand{\algorithmicrequire}{\textbf{Input:}}
\renewcommand{\algorithmcfname}{Algorithm}
\renewcommand{\algorithmicensure}{\textbf{Output:}}
\begin{algorithm}[!]
	\caption{SAP-CTI}
	\begin{algorithmic}[1]
		\Require Customer holds the private training data $\mathcal{D}:=\{(\bm{x}_{i},y_{i})|i=1,2, \dots,|\mathcal{D}|\}$. Vendor holds the pre-trained model $\bm{w}$. Set the privacy parameter $\eta$,  the number of epochs $E$,  and batch size $b$ appropriately.
		\renewcommand{\algorithmicrequire}{\textbf{Initialization:}}
		\Require
		\State {\bfseries at Vendor} 
		\State \quad Load the pre-trained model $\bm{w}$ and initialize the fine-tuning module. 
            \State \quad Split the entire model $\bm{w}$ into a top model $\bm{w}_t$ and a bottom model $\bm{w}_b$, where $\bm{w}_b$ is a frozen embedding layer.
            \State \quad Send the bottom model $\bm{w}_b$ to Customer.
		\State {\bfseries end at Vendor}
            \State {\bfseries at Customer} 
            \State \quad For each sample $\bm{x}_i$,
            compute the text representation $\phi(\bm{x}_i)=[\phi(x_{i}^{1}), \phi(x_{i}^{2}), \dots,\phi(x_{i}^{n})]$.
            \State \quad Compute the utility importance of each token in the vocabulary to each class. 
            \State \quad Select the top-$k$ tokens with the largest UI values as contributing tokens for each class.
            \State \quad For each non-contributing token, add  independent random noise to the corresponding embedding vector $\hat{\phi}(x_{i}^{j}) = \phi(x_{i}^{j}) + \bm{n}$, where the probability density of the noise satisfies $p(\bm{n}) \propto \exp (-\eta\|\bm{n}\|)$.
            \State \quad Map each perturbed vector to its nearest neighbor in the embedding space according to \eqref{eqn4}.
            \State \quad Send the perturbed text representations $\{\bar{\phi}(\bm{x_{i}})|i=1,2, \dots,|\mathcal{D}|\}$ to Vendor.
            \State {\bfseries end at Customer}
		\renewcommand{\algorithmicrequire}{\textbf{Fine-tuning:}}
		\Require
		\State {\bfseries at Vendor} 
            \State \quad {\bfseries for} each epoch $e=1,2,\dots,E$ {\bfseries do}
            \State \quad \quad Split $\{\bar{\phi}(\bm{x_{i}})|i=1,2, \dots,|\mathcal{D}|\}$ into batches of size $b$ and get the batch set $\mathcal{B}$.
            \State \quad \quad {\bfseries for} each batch $\bm{B} \in \mathcal{B}$  {\bfseries do}      
            \State \quad \quad \quad Perform forward propagation in the top model to compute the output $g(\bm{w}_t;\bm{B})$.
            \State \quad \quad \quad Send the output $g(\bm{w}_t;\bm{B})$ to Customer.
            \State \quad \quad \quad {\bfseries at Customer} 
            \State \quad \quad \quad \quad Compute the loss value $l=\mathcal{L}\left(g(\bm{w}_t;\bm{B});\bm{y}\right)$ and the gradient of the output layer $\frac{\partial l}{\partial g}$.
            \State \quad \quad \quad \quad Send the gradient $\frac{\partial l}{\partial g}$ to Vendor.
            \State \quad \quad \quad {\bfseries end at Customer}
            \State \quad \quad \quad Compute the gradient of the fine-tuning module and update it.
            \State \quad \quad {\bfseries end for}
            \State \quad {\bfseries end for}
            \State \quad Incorporate the fine-tuning module into the PLM.
		\State {\bfseries end at Vendor}
		\Ensure Fine-tuned large language model.		 
	\end{algorithmic}
\end{algorithm}

\subsection{Specific Implementation}
In specific implementation, the SAP framework can utilize a variety of existing PEFT algorithms to reduce computation cost.
Algorithm 1 provides an implementation example of the SAP framework, where the bottom model is the frozen embedding layer, the privatization mechanism \cite{feyisetan2020privacy} is adopted to guarantee $\eta d\chi$-privacy, and the CTI method is used to improve the utility-privacy trade-off. 
During the inference phase, the customer applies the same privatization mechanism to perturb the text representations to be transmitted and receives the model output returned by the vendor.

\section{Empirical Experiments}
In this section, we conduct extensive experiments to evaluate the effectiveness of the SAP framework. First, we give the experimental results of Algorithm 1 on multiple datasets in Section 5.2, including privacy attacks and performance comparisons. Subsequently, in Section 5.3 and Section 5.4, we delve into the impact of unfrozen bottom model and split position on the SAP framework, respectively. 

\subsection{Experiment Settings}
To begin with, let us introduce the experiment settings, including the pre-trained model, datasets, attack methods, and implementation details.

\textbf{Model and Datasets.}
In the experiments, we utilize the Roberta-Large \cite{liu2019roberta} published by Huggingface\footnote{ https://huggingface.co/roberta-large} as the pre-trained model, which consists of a total of 355 million parameters. We evaluate the SAP framework on the Financial Phrasebank (FP) \cite{malo2014good}, the Blog dataset used in \cite{lyu2020differentially}, as well as Stanford Sentiment Treebank (SST) and Microsoft Research Paraphrase Corpus (MRPC) datasets from the GLUE benchmark\footnote{https://gluebenchmark.com}. Detailed descriptions of these datasets are provided in Table 1. In the Blog dataset, each blog post also has the corresponding gender annotation of the author, which will be used to carry out the attribute inference attack.

\begin{table*}[!]
	\caption{Description of four datasets.}
	\centering
	\renewcommand\tabcolsep{7pt} 
	\renewcommand\arraystretch{1.38}
	\begin{tabular}{ccccc} 
		\hline 
		\hline
		Dataset & Task & \#Train & \#Dev & \#Test\\
            \hline
            FP &  sentiment analysis & 1808 & 226 & 226 \\
            Blog &  topic classiﬁcation & 7098 & 887 & 887 \\
            SST  &  sentiment analysis  & 66675 & 674 & 872 \\
            MRPC &  semantic equivalence judgment & 3301 & 367 & 408 \\
		\hline
		\hline
	\end{tabular}
\end{table*}

\begin{table*}[!]
	\caption{Performance ($\%$) of centralized fine-tuning on four datasets.}
	\centering
	\renewcommand\tabcolsep{7pt} 
	\renewcommand\arraystretch{1.38}
	\begin{tabular}{ccccc} 
		\hline 
		\hline
		Dataset & FP & Blog & SST & MRPC\\
            \hline
            Accuracy & $98.75$ & $96.71$ & $95.89$ & $89.42$  \\
		\hline
		\hline
	\end{tabular}
\end{table*}

\textbf{Attack Methods.} Following the work \cite{song2020information}, simulated attacks are employed to investigate the capability of the SAP framework to protect customer's data privacy. To maximize the attacker's abilities, we consider the white-box setting, assuming that the attacker has the same perspective as the vendor and can access the perturbed text representations transmitted by the customer as well as the initial parameters of the bottom model. The attack methods used are listed as follows:
\begin{itemize}[leftmargin=0.5cm]
\setlength{\itemsep}{0pt}
\setlength{\parsep}{0pt}
\setlength{\parskip}{0pt}
\item Embedding inversion attack (EIA) is a token-level attack whose goal is to recover the original input text from the perturbed text representations. Specifically, for the case where the bottom model only has the embedding layer, the nearest neighbor of each perturbed embedding is searched in the embedding space as a prediction of the original token \cite{qu2021natural}. For the case where the bottom model  contains more layers, a complex optimization-based attack method proposed in \cite{song2020information} is employed. For each input sample, the method iteratively optimizes the word selection vectors by minimizing the distance between the predicted text's representations and the observed representations.

\item Since the text representations still contain rich semantic information, attackers can launch the attribute inference attack (AIA) to infer sensitive attributes of users from them. In this paper, it is assumed that the attacker can obtain privacy attribute labels of some samples, such as the author's gender in the Blog dataset. The privacy inference problem is then treated as a downstream task, and a classifier is trained using the text representations of these samples and the corresponding privacy labels \cite{song2020information}. After training, the attacker can obtain predictions for the privacy attributes of other samples by feeding their representations into the classifier.
\end{itemize}  

\textbf{Implementation Details.}	
Our experiments are all implemented based on the Transformers library and PEFT library of Huggingface. Specifically, the LoRA \cite{hu2021lora} method is adopted to fine-tune the Roberta-Large model, and the AdamW optimizer with a linear 
learning rate scheduler is used during fine-tuning, where the initial learning rate is set to 3e-4. Following the work \cite{li2023privacy}, we use classification accuracy and empirical privacy as metrics to evaluate utility performance and privacy protection capability, where empirical privacy is defined as $1-X$ and $X$ represents the attack success rate. As a baseline, Table 2 presents the performance of centralized fine-tuning with the same settings.
The following experiments are performed on a server with one NVIDIA Tesla V100 GPU.

\subsection{Results of Algorithm 1}
\begin{figure*}[!]
    \centering
    \subfigure[Financial Phrasebank]{
        \begin{minipage}[t]{0.5\linewidth}
            \centering
            \includegraphics[width=3.2in]{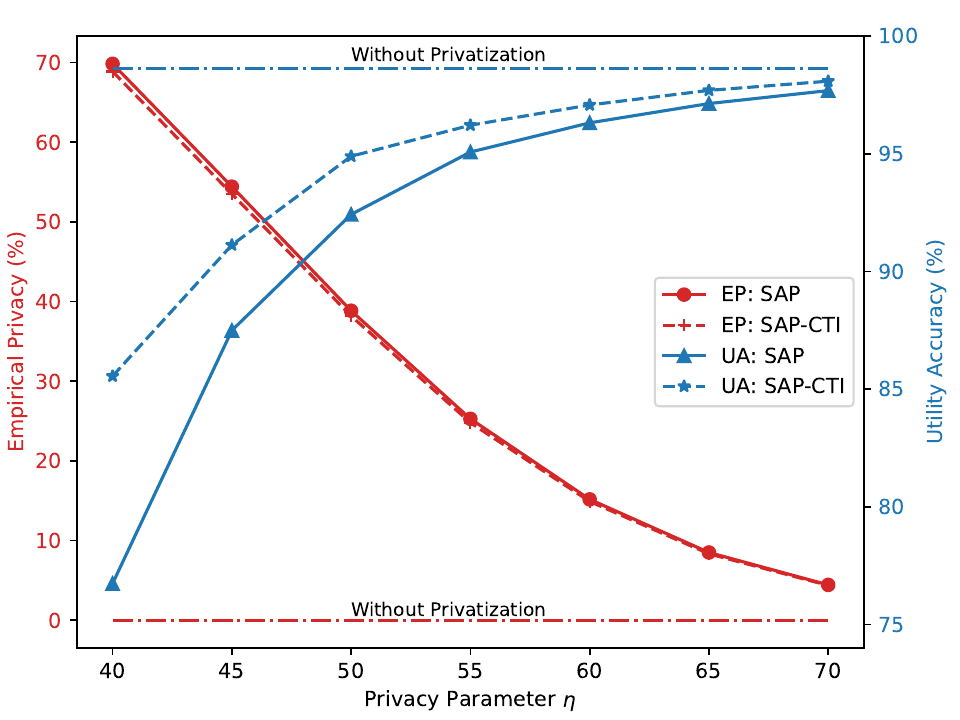}
        \end{minipage}
    }%
    \subfigure[Blog]{
        \begin{minipage}[t]{0.5\linewidth}
            \centering
            \includegraphics[width=3.2in]{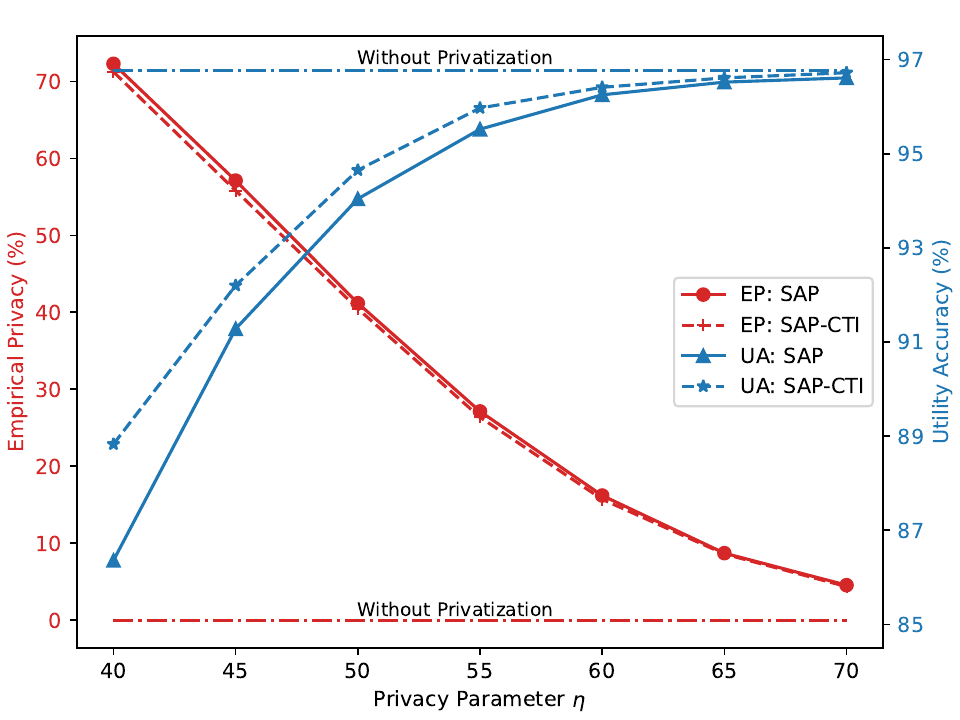}
        \end{minipage}
    }%
    
    \subfigure[Stanford Sentiment Treebank]{
        \begin{minipage}[t]{0.5\linewidth}
            \centering
            \includegraphics[width=3.2in]{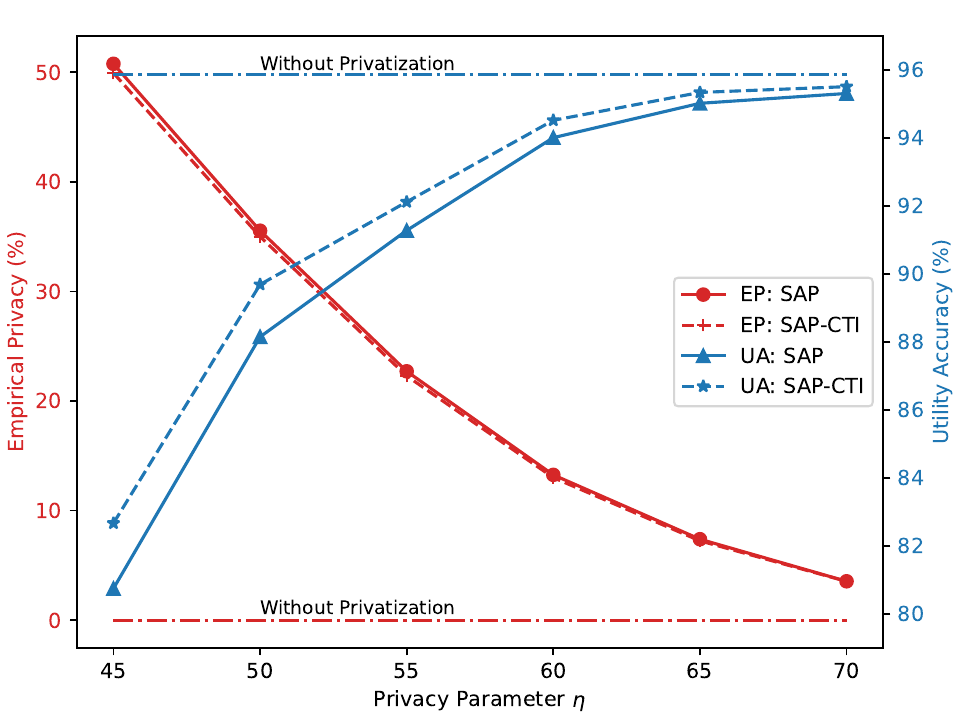}
        \end{minipage}%
    }%
    \subfigure[Microsoft Research Paraphrase Corpus]{
        \begin{minipage}[t]{0.5\linewidth}
            \centering
            \includegraphics[width=3.2in]{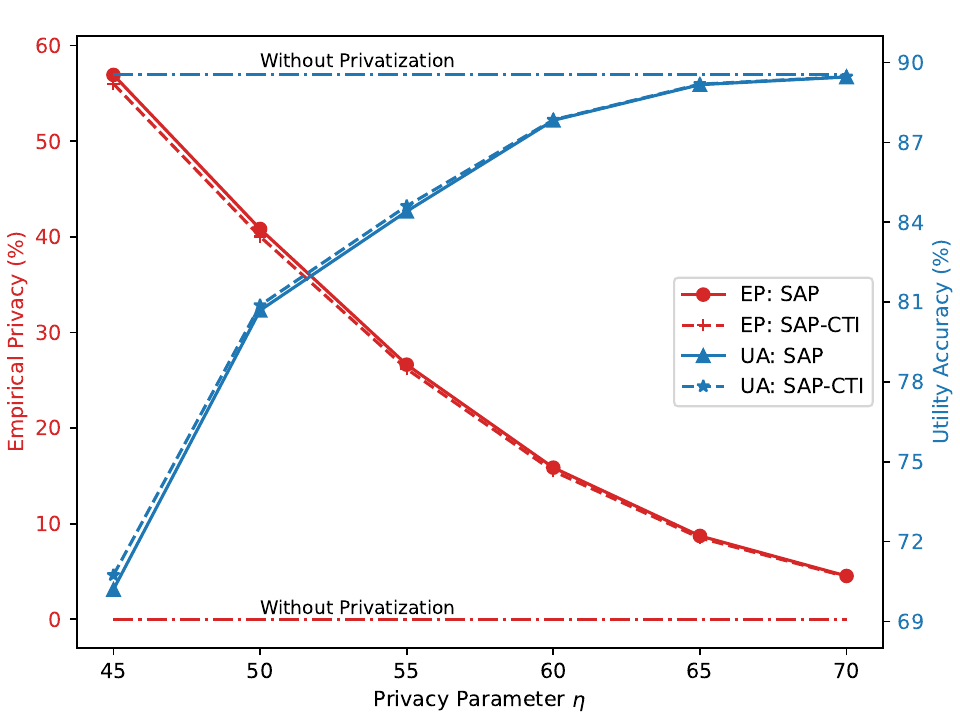}
        \end{minipage}
    }%
    \centering
    \caption{Impact of the privacy parameter $\eta$ on the empirical privacy (EP) against EIA and utility accuracy (UA).} 
\end{figure*}

First, we investigate the impact of the privacy parameter $\eta$ on the privacy protection capability and utility performance of the SAP framework, where the bottom model is a frozen embedding layer. For the SAP-CTI algorithm, the value of $k$ is set to the maximum value that ensures the number of contributing tokens does not exceed $1\%$ of the total tokens in the training dataset.
From the experimental results in Figure 2, it can be observed that when the bottom model is a frozen embedding layer, model split without text privatization does not result in performance loss compared with centralized fine-tuning. However, if embedding vectors are released without privatization, the attacker can easily recover the input text with an attack success rate of up to $100\%$.
By adding perturbations to text representations to guarantee $\eta d\chi$-privacy, the privacy protection capability of the SAP framework is strengthened. It can be seen that as the privacy parameter $\eta$ decreases, we obtain better empirical privacy against EIA, but at the same time, the utility accuracy keeps decreasing. In other words, the SAP framework involves a trade-off between utility and privacy. For example, on the FP dataset, SAP improves the empirical privacy to $38.85\%$ at the cost of $6.17\%$ performance degradation when $\eta$ is set to $50$. 

Furthermore, the experimental results on the FP, Blog, and SST datasets demonstrate that the CTI method effectively improves the trade-off between utility and privacy of the SAP framework. By reducing the perturbations on contributing tokens that account for less than $1\%$, the CTI method can significantly improve utility accuracy while maintaining empirical privacy almost unchanged. Specifically, on the FP dataset with $\eta$ set to $50$, the SAP-CTI algorithm can achieve the empirical privacy of $38.29\%$ with only $3.67\%$ performance loss. 
However, the CTI method cannot yield significant improvements on the MRPC dataset. This is because the CTI method only performs correlation analysis at the token level, while for tasks like semantic equivalence judgment, the class label of each sample is less relevant to individual tokens and more relevant to the overall semantics of the sentence. This is an inherent limitation of the CTI method.

\begin{figure}[t]
\centering
\includegraphics[width=3.2in]{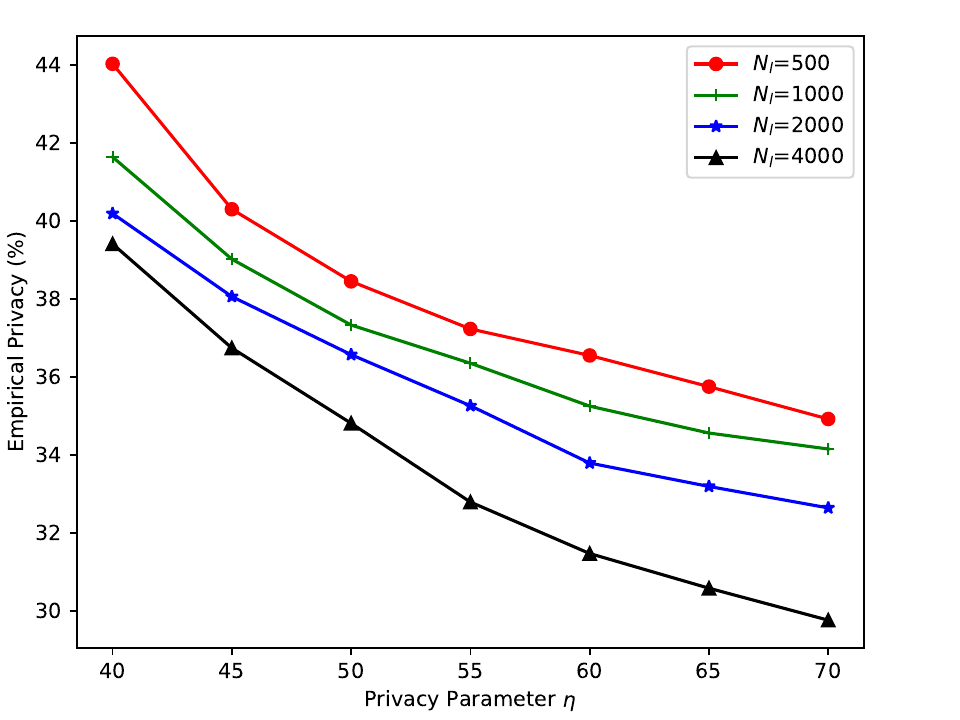}
\caption{Results of the SAP framework defending against AIA on the Blog dataset.}\label{fig3}
\end{figure}

Figure 3 presents the results of the SAP framework defending against AIA on the Blog dataset with different numbers of labeled data $N_l$. Specifically, the attacker fine-tunes the Roberta model using some auxiliary gender labels along with the corresponding text representations sent by the customer to infer the gender labels of other samples.
The results indicate that the attack success rate of attacker is positively related to the amount of labeled data it possesses.
By reducing the privacy parameter, the SAP framework becomes more capable of defending against attribute inference, which is consistent with the results of EIA.


\subsection{Impact of Unfrozen Bottom Model}
\begin{table*}[!]
	\caption{Empirical privacy (against EIA) and utility accuracy comparison on the SST dataset when the bottom model is a frozen or unfrozen embedding layer.}
	\centering
	\renewcommand\arraystretch{1.35}
	\begin{tabular}{c|c|cccccc} 
		\hline 
		\hline
		\multirow{2}{*}{Bottom Model}  & \multirow{2}{*}{Metric} &  \multicolumn{6}{c}{Privacy Parameter $\eta$}   \\
		\cline{3-8}  
		& & 45 & 50 & 55 & 60 & 65 & 70        \\    
		\hline    
		\multirow{2}{*}{Frozen Embedding Layer}  &  EP  &  50.77 & 35.54 & 22.71 & 13.26 & 7.39 & 3.55	          \\ 
            \cline{2-8} 
		& UA  & 80.75 & 88.15 & 91.28 & 94.01 & 95.02 & 95.31         \\ 
            \hline    
		\multirow{2}{*}{Unfrozen Embedding Layer}  &  EP & 18.47 & 7.51 & 2.45 & 0.93 & 0.21 & 0    \\ 
            \cline{2-8}
		& UA & 81.12 & 88.23 & 92.13 &  94.08 & 95.29 & 95.54    \\ 
		\hline
		\hline
	\end{tabular}
\end{table*}

Next, we focus on the privacy protection capability and utility performance of the SAP framework when training the bottom model and top model together. When the bottom model is unfrozen, the customer needs to send a batch of sample representations in each iteration. If the training is conducted for $E$ epochs, the attacker will have access to $E$ representations for each sample,  and it can independently use each representation to launch the EIA. The input tokens successfully inferred by the attack are the union of the results from EIA conducted $E$ times. The results in Table 3 indicate that when the bottom model consists only of an embedding layer, joint training of the bottom model has a minor impact on utility accuracy but results in a significant decrease in empirical privacy. The reason is that the parameters of the embedding layer change slightly after fine-tuning, and the attacker can still use the initial parameters to perform nearest neighbor search. Therefore, multiple observations of the same sample will significantly increase the attack success rate.

\begin{figure}[t]
\centering
\includegraphics[width=3.2in]{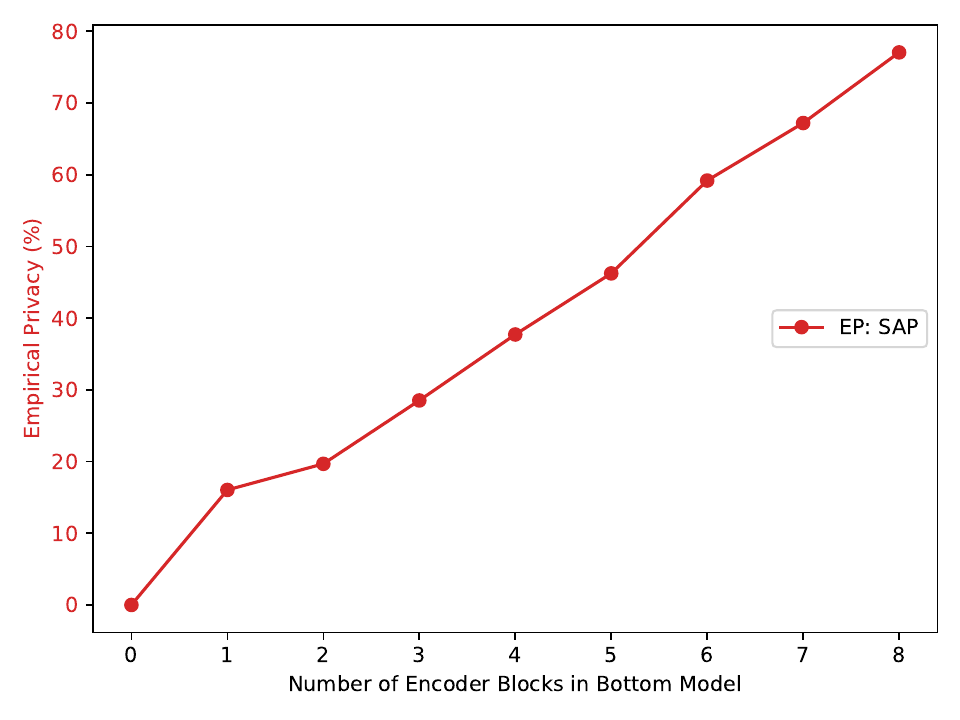}
\caption{Empirical privacy against EIA of SAP framework (without privatization) with different split positions on the SST dataset.}\label{fig4}
\end{figure}

\subsection{Impact of Split Position}
In the SAP framework, the split position of the PLM is an important option. The Roberta model comprises a total of 24 encoder blocks. In the experiment, we split the model after the 1st to 8th encoder block and compare them with the case where the bottom model only has the embedding layer. Figure 4 gives the impact of different split positions on the empirical privacy against EIA of the SAP framework without text privatization. The results show that as the number of encoder blocks in the bottom model increases, it becomes increasingly difficult for an attacker to infer the input text from the representations transmitted by the customer. Even without text privatization, the empirical privacy reaches about $80\%$ when there are 8 encoder blocks in the bottom model. 

Furthermore, we delve into the privacy protection capability and utility performance of the SAP framework with different split positions and different privacy parameter settings. Compared with centralized fine-tuning, the results in the last column of Table 4 indicate that as the number of layers included in the bottom model increases, the utility accuracy of the SAP framework without privatization slightly decreases. This is reasonable because the number of layers that can be fine-tuned decreases accordingly when the bottom model is frozen. 
In addition, we can observe that by applying text privatization and reducing the privacy parameter, empirical privacy is further strengthened, but at the same time, the utility accuracy also decreases, which is consistent with the results in Figure 2.

From the above experimental results, we can conclude that if the customer has some computation resources, moderately splitting more layers of the PLM into the bottom model helps achieve a better balance between model utility and data privacy.


\begin{table*}[!]
	\caption{Empirical privacy (against EIA) and utility accuracy comparison of SAP framework with different split positions and different privacy parameter settings on the SST dataset, where ``None'' represents the case without privatization. }
	\centering
	\renewcommand\arraystretch{1.35}
	\begin{tabular}{c|c|ccccccc} 
		\hline 
		\hline
		\multirow{2}{*}{Bottom Model (Frozen)}  & \multirow{2}{*}{Metric} &  \multicolumn{7}{c}{Privacy Parameter $\eta$}   \\
		\cline{3-9}  
		& & 25 & 30 & 35 & 40 & 45 & 50  & None      \\    
		\hline    
		\multirow{2}{*}{Embedding and 2 encoder blocks }  &  EP & 23.97  & 22.35 & 21.61  & 21.06 & 20.64 & 20.32 & 19.68          \\  	
            \cline{2-9} 
		& UA & 91.97  & 94.54 & 95.32 & 95.52 & 95.61 & 95.76  & 95.84     \\ 
  
            \hline    
		\multirow{2}{*}{Embedding and 4 encoder blocks}  &  EP & 41.84  & 40.29 & 39.58 & 39.23 &  38.81& 38.47 &	37.80  \\ 
            \cline{2-9}
		& UA & 91.86  & 94.31 & 94.93 & 95.02 & 95.24 & 95.40	& 95.72\\ 
  
        \hline    
		\multirow{2}{*}{Embedding and 6 encoder blocks }  &  EP & 70.14  & 65.22 & 62.71 & 61.97  & 61.29 & 60.84 & 59.19         \\ 
            \cline{2-9} 
		& UA & 86.09  & 92.87 & 94.01 & 94.75 & 95.14 & 	95.34 &   95.53  \\ 
            \hline           					
		\multirow{2}{*}{Embedding and 8 encoder blocks}  &  EP & 84.78  & 81.91 & 80.36 & 79.47 & 78.85 & 78.29 &	77.06 \\ 
            \cline{2-9}
		& UA & 83.89  & 90.06 & 92.81 & 93.79 & 94.37  & 94.72 & 95.21	 \\ 					
		\hline
		\hline
	\end{tabular}
\end{table*}


\section{Conclusion}
In this paper, we have considered the model privacy and data privacy issues in LLM customization and proposed a privacy-preserving federated fine-tuning framework called SAP. By splitting the PLM into a top model on the vendor and a bottom model on the customer, and applying text privatization techniques to perturb the text representations transmitted by the customer, the proposed SAP framework can achieve a good balance between protecting model privacy and data privacy while maintaining competitive performance. Moreover, SAP is a flexible framework capable of adapting to various requirements and scenarios in LLM customization services. 
For cases where the customer's computation resources are scarce, it is more feasible to adopt a solution where the bottom model is a frozen embedding layer. Experimental results indicate that it enhances the empirical privacy by $36\%$ at the cost of $6\%$ model performance degradation on the SST dataset. 
For cases where the customer has relatively abundant computation resources, it is advisable to adopt a solution where the bottom model contains 6 encoder blocks (constituting a quarter of the PLM). Experimental results indicate that it can enhance the empirical privacy by $62\%$ at the cost of $1\%$ model performance degradation on the SST dataset.


\bibliographystyle{unsrt}  
\bibliography{references}

\end{document}